%% file: dialrec.tex
\PassOptionsToPackage{usenames,dvipsnames,table}{xcolor}
\documentclass[11pt,a4paper]{article}
\usepackage{emnlp-ijcnlp-2019}
\usepackage{times}
\usepackage{latexsym}
\usepackage{url}
\usepackage{amssymb}
\usepackage{graphicx}
\usepackage{times}
\usepackage{helvet}
\usepackage{courier}
\usepackage{balance}  
\usepackage{mdwlist}
\usepackage{graphics}
\usepackage{color}
\usepackage{rotating}
\usepackage{booktabs}
\usepackage{epsfig}
\usepackage{alltt}
\usepackage{moreverb}
\usepackage{fancyvrb}
\usepackage{enumerate}
\usepackage{xspace}
\usepackage{relsize}
\usepackage{amsmath}
\usepackage{amsfonts}
\usepackage{txfonts}
\usepackage{tabularx}
\usepackage{caption}
\usepackage{rotating}
\usepackage{multirow}
\usepackage{booktabs}
\usepackage{float}
\usepackage{array}
\usepackage[toc,page]{appendix}
\usepackage{enumitem}
\usepackage{subfig}
\usepackage{arydshln}
\usepackage{setspace}
\usepackage{rotating}
\usepackage{pbox}
\usepackage{todonotes}
\usepackage{bm}
\usepackage{algorithm}
\usepackage{tipa} 
\usepackage[export]{adjustbox}
\usepackage{makecell}

\newcommand{\method}{\textsc{GoRecDial}\xspace}

\usepackage{etoolbox}
\newtoggle{draft}
\toggletrue{draft}
\iftoggle{draft}{
\newcommand{\dk}[1]{\textcolor{Maroon}{[#1 \textsc{--dongyeop}]}}
\newcommand{\ed}[1]{\textcolor{Red}{[#1 \textsc{--ed}]}}
\newcommand{\jason}[1]{\textcolor{BurntOrange}{[#1 \textsc{--jason}]}}
\newcommand{\anusha}[1]{\textcolor{Orange}{[#1 \textsc{--anusha}]}}
\newcommand{\ylan}[1]{\textcolor{ForestGreen}{[#1 \textsc{--ylan}]}}
}{
\newcommand{\dk}[1]{}
\newcommand{\ed}[1]{}
\newcommand{\jason}[1]{}
\newcommand{\anusha}[1]{}
\newcommand{\ylan}[1]{}
}

\aclfinalcopy 

\begin{document}

\title{Recommendation as a Communication Game:\\
Self-Supervised Bot-Play for Goal-oriented Dialogue}

\author{
\makecell{Dongyeop Kang$^\heartsuit$ \quad 
Anusha Balakrishnan$^\clubsuit$ \quad  Pararth Shah$^\clubsuit$ \,  \\
Paul Crook$^\clubsuit$ \quad  Y-Lan Boureau$^\clubsuit$ \quad Jason Weston$^\clubsuit$}\\
$^\heartsuit$Carnegie Mellon University, \quad $^\clubsuit$Facebook AI\\
{\tt dongyeok@cs.cmu.edu   } {\tt anusha.bala28@gmail.com}\\
\quad {\tt $\{$pararths,pacrook,ylan,jase$\}$@fb.com}
}

\maketitle

\input{00abstract.tex}

\input{01introduction.tex}

\input{02data.tex}

\input{03models.tex}

\input{04experiment.tex}

\input{06related.tex}

\input{05discussion.tex}

\section*{Acknowledgements}
We thank Eduard Hovy, Alan W Black, Dan Jurafsky, Alan Ritter, and anonymous reviewers for their helpful comments.

\bibliographystyle{acl_natbib}
\bibliography{dialrec}

\input{appendix}

\end{document}

%% file: 00abstract.tex
\begin{abstract}

Traditional recommendation systems produce static rather than interactive recommendations invariant to a user's specific requests, clarifications, or
current mood, and can
suffer from the cold-start problem if their tastes are unknown. 
These issues
can be alleviated by treating recommendation as 
an interactive dialogue task instead,
where an expert recommender can  sequentially ask 
about someone's preferences, react to their requests, and recommend more appropriate items. 
In this work, we collect a goal-driven recommendation dialogue dataset (\method), which consists of 9,125 dialogue games and 81,260 conversation turns between pairs of human workers recommending movies to each other.
The task is specifically designed as a cooperative game between two players working towards a quantifiable common goal.
We leverage the dataset to
develop an end-to-end dialogue system that can simultaneously converse and recommend.
Models are first trained to imitate the behavior of human players without considering the task goal itself (supervised training).
We then fine-tune our models on simulated bot-bot conversations between two paired pre-trained models (bot-play), in order to achieve the dialogue goal. 
Our experiments show that models fine-tuned with bot-play learn improved dialogue strategies, reach the dialogue goal more often when paired with a human, and are rated as more consistent by humans compared to models trained without bot-play.
The dataset and code are publicly available through the ParlAI framework\footnote{\url{https://github.com/facebookresearch/ParlAI}}.
\end{abstract}

%% file: 01introduction.tex
\section{Introduction}\label{sec:intro_dialrec}

Traditional recommendation systems factorize users' historical data (i.e., ratings on movies) to extract common preference patterns
\cite{Koren2009MatrixFT,He2017NeuralCF}. However, besides making it difficult to accommodate new users because of the {\em cold-start problem}, relying on aggregated history makes  these systems static, and prevents users from making specific requests, or exploring a temporary interest.
For example, a user who usually likes horror movies, but is in the mood for a fantasy movie, has no way to indicate their preference to the system, and would likely get a recommendation that is not useful. Further, they cannot iterate upon initial recommendations with clarifications or modified requests, all of which are best specified in natural language.

\begin{figure}[t]
\vspace{0mm}
\centering
{
\includegraphics[trim=0cm 8.2cm 14.1cm 0cm,clip,width=.99\linewidth]{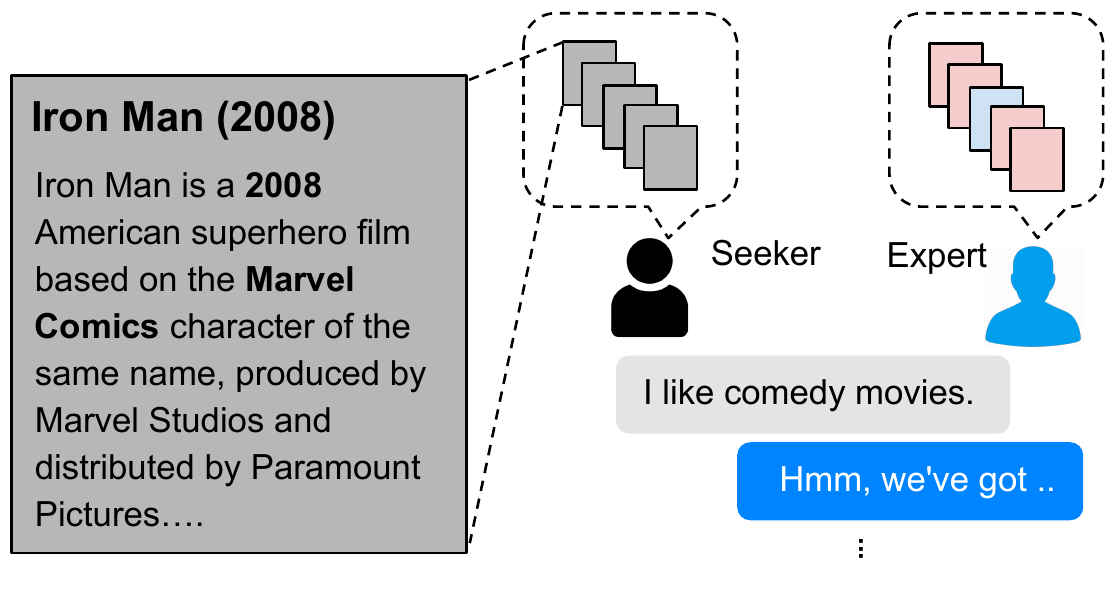}
}\vspace{-2mm}
\caption{\label{fig:amt_setup} Recommendation as a dialogue game. We collect 81,260 recommendation utterances between pairs of human players (experts and seekers) with a collaborative goal: the expert must recommend the correct (blue) movie, avoiding incorrect (red) ones, and the seeker must accept it. A chatbot is then trained to play the expert in the game. 
}
\vspace{0mm}
\end{figure}

Recommending through dialogue interactions \citep{Reschke2013GeneratingRD,Wrnestl2004ModelingAD} offers a promising solution to these problems, and  recent work by \citet{li2018towards} explores this approach in detail. However, the dataset introduced in that work does not capture higher-level strategic behaviors that can impact the quality of the recommendation made (for example, it may be better to elicit user preferences first, before making a recommendation). This makes it difficult for models trained on this data to learn optimal recommendation strategies. Additionally, the recommendations are not grounded in real observed movie preferences, which may make trained models less consistent with actual users. This paper aims to provide \textit{goal-driven recommendation dialogues grounded in real-world data}. 
We collect a corpus of goal-driven dialogues grounded in real user movie preferences through a carefully designed gamified setup (see Figure \ref{fig:amt_setup}) and show that models trained with that corpus can learn a successful recommendation dialogue strategy. The training is conducted in two stages: first, a {\em supervised} phase that trains the model to mimic human behavior on the task; second, a {\em bot-play} phase that improves the goal-directed strategy of the model.


The contribution of this work is thus twofold. (1) We provide the first (to the best of our knowledge) large-scale goal-driven recommendation dialogue dataset with specific goals and reward signals,  grounded in a real-world knowledge base. (2) We propose a two-stage recommendation strategy learning framework and empirically validate that it leads to better recommendation conversation strategies.


%% file: 02data.tex
\begin{figure*}[ht]
\vspace{0mm}
\centering
{
\includegraphics[trim=0.2cm 3.0cm 4.1cm 0cm,clip,width=.995\linewidth]{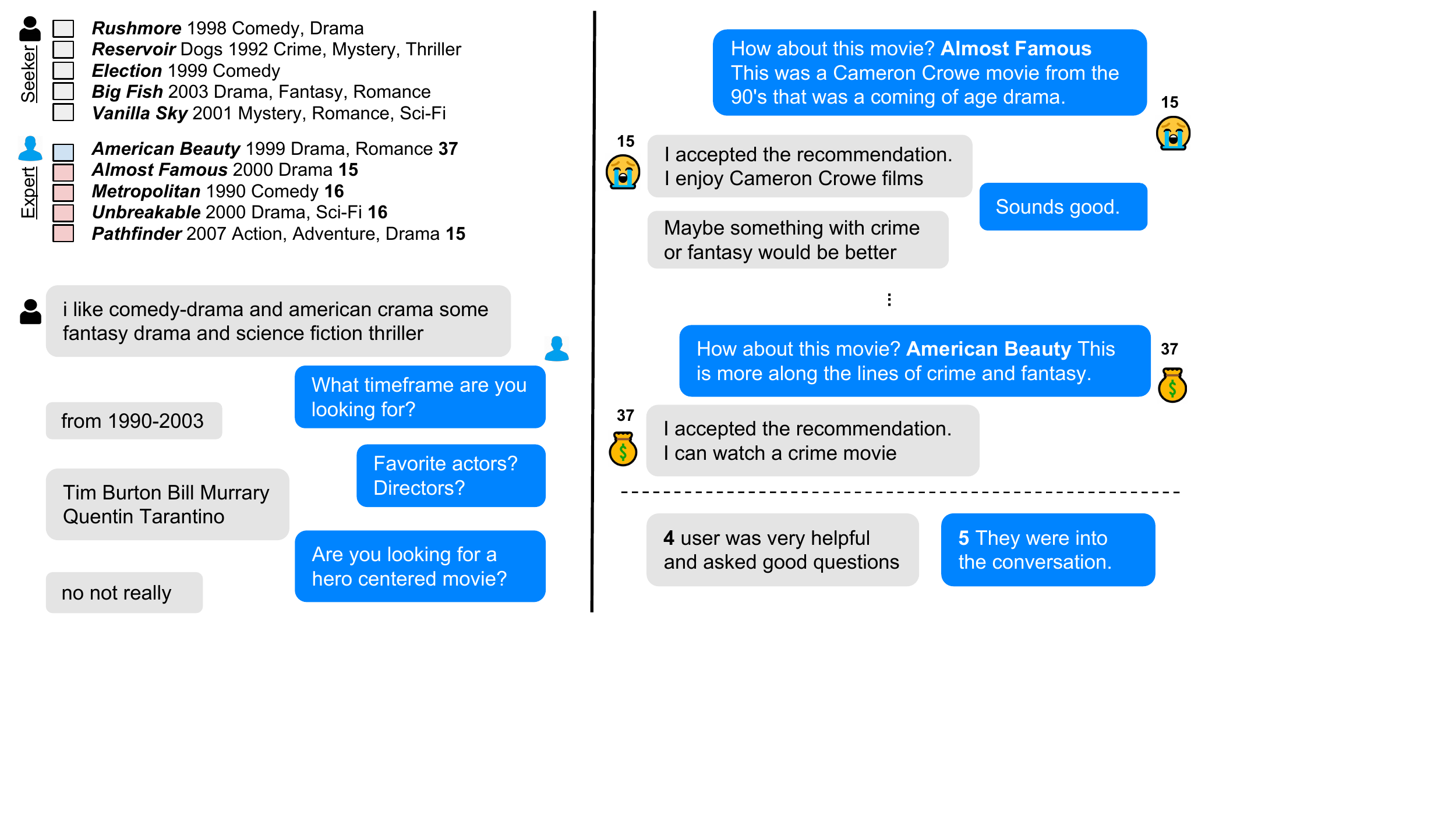}
}
\vspace{0mm}
\caption{\label{fig:example_dialog_long}An example dialogue from our dataset of movie recommendation between two human workers: seeker (grey) and expert (blue). The goal is for the expert to find and recommend the correct movie (light blue) out of incorrect movies (light red) which is similar to the seeker movies. Best viewed in color.
}
\vspace{0mm}
\end{figure*}

\section{Recommendation Dialogue Task Design}\label{sec:dataset_dialrec}


In this section, we first describe the motivation and design of the dialogue-based recommendation game that we created. We then describe the data collection environment and present detailed dataset statistics.

\subsection{Dialogue Game: Expert and Seeker}

The game is set up as a conversation between a {\em seeker} looking for a movie recommendation, and an {\em expert} recommending movies to the seeker.
Figure \ref{fig:example_dialog_long} shows an example movie recommendation dialogue between two-paired human workers on Amazon Mechanical Turk.

\paragraph{Game Setting.}
Each worker is given a set of five movies\footnote{We deliberately restricted the set of movies to make the task more tractable. 
One may argue that the expert can simply ask these candidates one by one (at the cost of low engagingness). However, this empirically doesn't happen: experts make on average only 1.16 incorrect movie recommendations.} with a description (first paragraph from the Wikipedia page for the movie) including important features such as director's name, year, and genre. The seeker's set represents their watching history (movies they are supposed to have liked) for the game's sake. The expert's set consists of candidate movies to choose from when making recommendations, among which only one is the \textit{correct} movie to recommend. The correct movie is chosen to be similar to the seeker's movie set (see Sec.~\ref{sec:set_prep}), while the other four movies are dissimilar. The expert is not told by the system which of the five movies is the correct one.
The expert's goal is to find the correct movie by chatting with the seeker and recommend it after a minimal number of dialogue turns. The seeker's goal is to accept or reject the recommendation from the expert based on whether they judge it to be similar to their set. The game ends when the expert has recommended the correct movie. The system then asks each player to rate the other for engagingness.


\paragraph{Justification.}
Players are asked to provide reasons for recommending, accepting, or rejecting a movie, so as to get insight into human recommendation strategies\footnote{Our model doesn't utilize this or the engagingness scores for learning, but these are potential future directions.}.


\paragraph{Gamification.}
Rewards and penalties are provided to players according to their decisions, to make the task more engaging and incentivize better strategies.
Bonus money is given if the expert recommends the correct movie, or if the seeker accepts the correct movie or rejects an incorrect one.


\subsection{Picking Expert and Seeker movie sets}
\label{sec:set_prep}
This section describes how movie sets are selected for experts and seekers.

\paragraph{Pool of movies}
To reflect movie preferences of real users, our dataset uses the MovieLens dataset\footnote{\url{https://grouplens.org/datasets/movielens/}}, comprising 27M ratings applied to 58K movies by 280K real users. 
We obtain descriptive text for each movie from Wikipedia\footnote{\url{https://dumps.wikimedia.org/}} (i.e., the first paragraph).
We also extract entity-level features (e.g., directors, actors, year) using the MovieWiki dataset \cite{miller2016key}
(See Figure \ref{fig:amt_setup}).
We filter out less frequent movies and user profiles (see Appendix), resulting in a set of 5,330 movies and 65,181 user profiles with their ratings.

\paragraph{Movie similarity metric}
In order to simulate a natural setting, the movies in the seeker's set should be similar to each other, and the correct movie should be similar to these, according to a metric that reflects coherent empirical preferences.
To compute such a metric, we train an embedding-driven recommendation model \cite{wu2018starspace}.\footnote{We also tried a classical matrix-factorization based recommendation model, which shows comparable performance to the embedding model.}
Each movie is represented as an embedding, which is trained so that embeddings of movies watched by the same user are close to each other.
The closeness metric between two movies is the cosine similarity of these trained embeddings. A movie is deemed close to a set of movies if its embedding is similar to the average of the movie embeddings in the set.


\begin{figure}[h!]
\vspace{0mm}
\centering
{
\includegraphics[trim=0cm 15.5cm 16.7cm 0cm,clip,width=.99\linewidth]{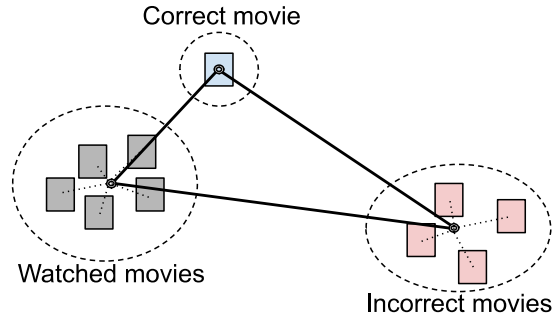}
}
\caption{\label{fig:movie_selection} Movie set selection: watched movies for seeker (grey) and correct (light blue) / incorrect (light red) movies for expert.
}
\vspace{0mm}
\end{figure}

\paragraph{Movie Set Selection}
Using these trained embeddings, we design seeker and expert sets based on the following criteria (See Figure \ref{fig:movie_selection}):
\begin{itemize}[noitemsep,topsep=0pt,leftmargin=*]
\item Seeker movies (grey) are a set of five movies which are close to each other, chosen from the set of all movies watched by a real user.
\item The correct movie (light blue) is close to the average of the five embeddings of the seeker set.
\item The expert's incorrect movies (light red) are far from the seeker set and the correct movie.
\end{itemize}
We filter out movie sets that are too difficult or easy for the recommendation task (see Appendix), and choose 10,000 pairs of seeker-expert movie sets at random.

\subsection{Data Collection}
For each dialogue game, a movie set is randomly chosen without duplication. 
We collect 
dialogues using ParlAI \cite{miller2017parlai} to interface with Amazon Mechanical Turk.
More details about data collection
are included in the Appendix.

\begin{table}[h]
\small
\centering
\begin{tabularx}{\linewidth}{@{}rc@{}}
\toprule
\multicolumn{2}{c}{\textbf{Dialogue statistics}} \\\midrule
Number of dialogues &\makecell{9,125} \\
Number of utterances & \makecell{170,904}\\
Number of unique utterances & \makecell{85,208}\\
Avg length of a dialogue & 23.0 \\
Avg duration (minutes) of a dialogue & 5.2\\
\toprule
\multicolumn{2}{c}{\textbf{Expert's utterance statistics}} \\\midrule
Avg utterance length & 8.40 \\ 
Unique tokens & 11,757\\
Unique utterances & 40,550 \\
\toprule
\multicolumn{2}{c}{\textbf{Seeker's utterance statistics} } \\\midrule
Avg utterance length & 8.47\\
Unique tokens & 10,766\\
Unique utterances & 45,196\\
\toprule
\multicolumn{2}{c}{\textbf{Action statistics}  (all scores are averaged)} \\\midrule
\# of correct/incorrect recs. by expert & 1.0 / 1.16 \\
\# of correct/incorrect decisions by seeker & 1.1 / 1.04\\
\toprule
\multicolumn{2}{c}{\textbf{Game statistics} (all scores are averaged)} \\\midrule
min/max movie scores & 12.3 / 46.0\\
correct/incorrect movies& 39.9 / 15.0 \\
real game score by expert/seeker & 61.3 / 50.8 \\
random game score by expert/seeker & 43.2 / 38.1 \\
\toprule
\multicolumn{2}{c}{\textbf{Engagingness statistics}  (all scores are averaged)} \\\midrule
engagingness score by expert/seeker & 4.3 / 4.4 \\
engagingness scores \& feedback collected & 18,308 \\
\bottomrule
\end{tabularx}
\caption{\label{tab:data} Data statistics. ``correct/incorrect'' in the action stats means that the expert recommends the correct/incorrect movie or the seeker correctly accepts/rejects the movie.}
\end{table}

Table \ref{tab:data} shows detailed statistics of our dataset regarding the movie sets, the annotated dialogues, actions made by expert and seeker, dialogue games, and engagingness feedback.  


The collected dialogues contain a wide variety of action sequences (recommendations and accept/reject decisions). Experts make an average of 1.16 incorrect recommendations, which indicates
a reasonable difficulty level.
Only $37.6\%$ of dialogue games end at first recommendation, and $19.0\%$ and $10.8\%$ at second and third recommendations, respectively. 


\begin{figure}[h!]
\vspace{-4mm}
\centering
{
\subfloat[Decision to speak (-1) \newline or recommend (+1)]{
\includegraphics[trim=0.4cm 5.2cm 8cm 1.0cm,clip,valign=b,width=.48\linewidth]{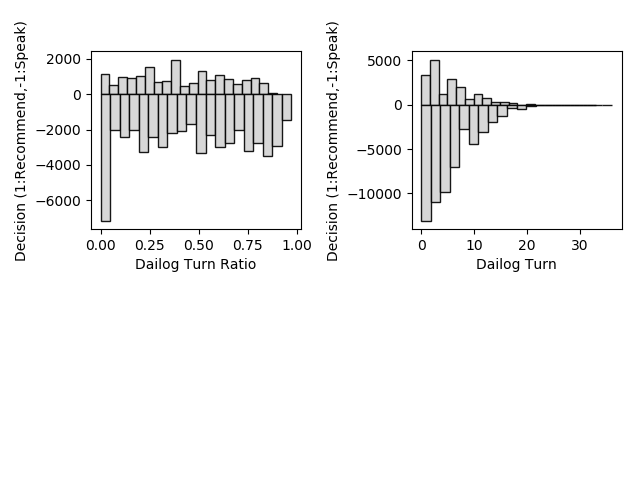}
}
\subfloat[Correct (+1) or incorrect (-1) recommendations]{
\includegraphics[trim=0.4cm 0.4cm 8.3cm 0.5cm,clip,valign=b,width=.455\linewidth]{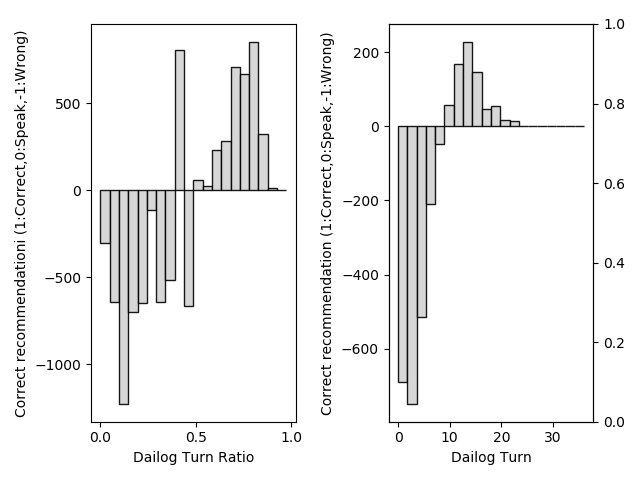}
}
}
\caption{\label{fig:analysis_dataset} Histogram distribution of (a) experts' decisions of whether to speak or recommend and (b) correct/incorrect recommendations over the normalized dialogue turns. 
}
\end{figure}

Figure \ref{fig:analysis_dataset} shows histogram distributions of (a) expert's decisions between speaking utterance and recommendation utterance and (b) correct and incorrect recommendations over the normalized turns of dialogue.
In (a), recommendations increasingly occur after a sufficient number of speaking utterances. 
In (b), incorrect recommendations are much more frequent earlier in the dialogue, while the opposite is true later on.

%% file: 03models.tex
\section{Our Approach}\label{sec:models_dialrec}

\begin{figure*}[h!]
\vspace{0mm}
\centering
{
\subfloat[Supervised multi-aspect learning]
{
\includegraphics[trim=0cm 6.3cm 13.2cm 0cm,clip,valign=t,width=.602\linewidth]{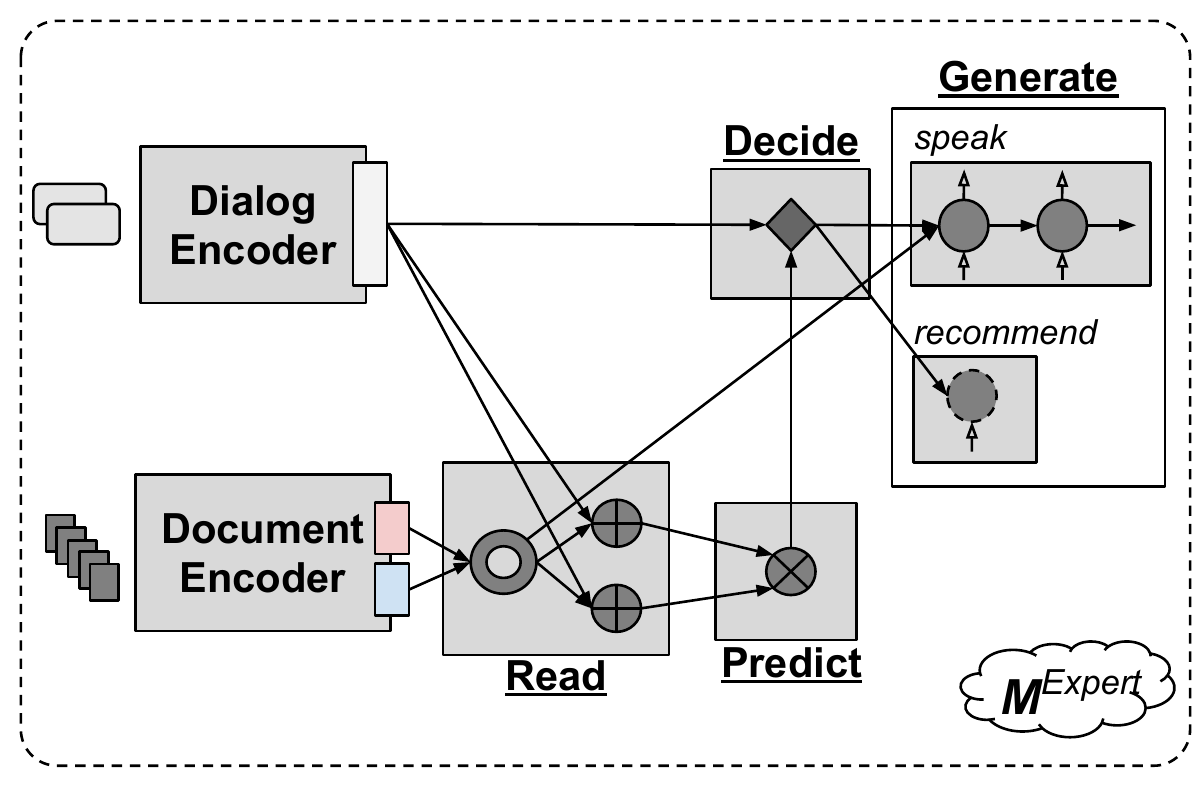}
}
\subfloat[Bot-play]
{
\includegraphics[trim=0cm 6.9cm 18.1cm 0cm,clip,valign=t,width=.392\linewidth]{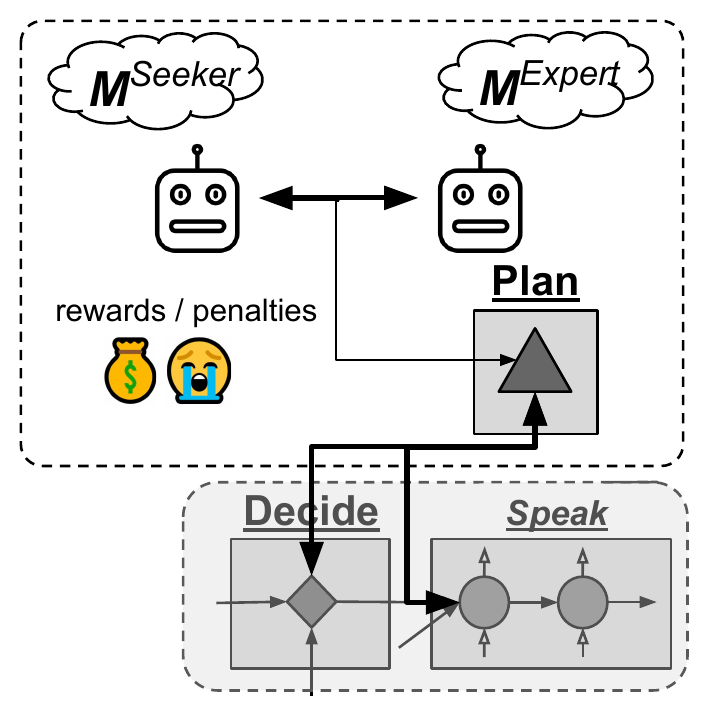}
}
}
\caption{\label{fig:models_dialrec} (a) Supervised learning of the expert model $\mathcal{M}^{expert}$ and (b) bot-play game between the expert $\mathcal{M}^{expert}$ and the seeker $\mathcal{M}^{seeker}$ models. 
The former imitates multiple aspects of humans' behaviors in the task, while the later fine-tunes the expert model w.r.t the game goal (i.e., recommending the correct movie).
}
\vspace{0mm}
\end{figure*}

In order to recommend the right movie in the role of the expert, a model needs to combine several
perceptual and decision skills. 
We propose to conduct learning in two stages (See Figure \ref{fig:models_dialrec}): \textit{supervised multi-aspect learning} and \textit{bot-play}.

\subsection{Supervised Multi-Aspect Learning}
The supervised stage of training the expert model combines three sources of supervision, corresponding to the three following subtasks: (1) \textsc{generate} dialogue utterances to speak with the seeker in a way that matches the utterances of the human speaker, (2) \textsc{predict}  the correct movie based on the dialogue history and the movie description representations, and (3) \textsc{decide} whether to recommend or speak in a way that matches the observed decision of the human expert.

Using an LSTM-based model \cite{hochreiter1997long},
 we represent the dialogue history context $h_t$ of utterances $x_1$ to $x_t$ as the average of LSTM representations of $x_1, \cdots, x_t$, and the description $m_k$ of the $k$-th movie as the average of the bag-of-word representations\footnote{We empirically found that BOW works better than other encoders such as LSTM in this case.} of its description sentences. Let $(x_{t+1},y,d_{t+1})$ denote the ground truth next utterance, correct movie index, and ground truth decision at time $t+1$, respectively. 
We cast the supervised problem as an end-to-end optimization of the following loss:
\begin{align}\label{eq:supervised}
\mathcal{L}_{sup} = 
\alpha\mathcal{L}_{gen} + \beta\mathcal{L}_{predict} + (1{-}\alpha{-}\beta)\mathcal{L}_{decide}, 
\end{align}
where $\alpha$ and $\beta$ are weight hyperparameters optimized over the validation set, and $\mathcal{L}_{predict},\mathcal{L}_{decide}, \mathcal{L}_{gen}$ are negative log-likelihoods of probability distributions matching each of the three subtasks:

\begin{align}\label{eq:sublosses}
\mathcal{L}_{gen} &= -\log p_{gen}(x_{t+1} | h_t, m_1, \cdots, m_K), \\ 
\mathcal{L}_{predict} &= -\log p(y | c_1, \cdots , c_K),\quad\text{where}\\
c_j  &=  h_t \boldsymbol{\cdot} m_j \quad\quad \text{for}~ j\in1..K,\\
\mathcal{L}_{decide} &= p_{MLP}(d_{t+1}|h_t,c_1, \cdots, c_K ),
\end{align}
with $p_{gen}$ the output distribution of an attentive seq2seq generative model \cite{Bahdanau2015NeuralMT}, $p$ a softmax distribution over dot products $h_t\cdot m_k$ that capture how aligned the dialogue history $h_t$ is with the description $m_k$ of the $k$-th movie, and $p_{MLP}$ the output distribution of a multi-layer perceptron predictor that takes $c_1, \cdots, c_K$ as inputs\footnote{We experimented with various other encoding functions, detailed in the Appendix.}.

\subsection{Bot-Play}
Motivated by the recent success of self-play in strategic games \cite{silver2017mastering,alphastarblog,OpenAIdota} and in  negotiation dialogues \cite{Lewis2017DealON}, we show in this section how we construct a reward function to perform bot-play between two bots in our setting, with the aim of developing a better expert dialogue agent for recommendation.



\paragraph{\textsc{Plan}} optimizes long-term policies of the various aspects over multiple turns of the dialogue game by maximizing game-specific rewards. 
We first pre-train expert and seeker models individually: the expert model $\mathcal{M}^{expert}\,(\theta) = \min_\theta \mathcal{L}_{sup} $ is pre-trained by minimizing the supervised loss in Eq \ref{eq:supervised}, and the seeker model $\mathcal{M}^{seeker}(\phi)$ is a retrieval-based model that retrieves seeker utterances from the training set based on cosine similarity of the preceding dialogue contexts encoded using the BERT pre-trained encoder\footnote{See Sec.~\ref{sec:eval_supervised} for details on BERT. We also experimented with sequence-to-sequence models for modeling the seeker but performance was much worse.}. $\theta$ and $\phi$ are model parameters of the expert and seeker model, respectively.
Then, we make them chat with each other, and fine-tune the expert model by maximizing its reward in the game (See Figure \ref{fig:models_dialrec}, Right). 

The dialogue game ends if the expert model recommends the correct movie, or a maximum dialogue length is reached\footnote{We restrict the maximum length of a dialogue to 20.
}, 
yielding $T$ turns of dialogue; $g = (x^{expert}_1,x^{seeker}_1..x^{expert}_T,x^{seeker}_T)$.
Let $T_{REC}$ the set of turns when the expert made a recommendation. We define the expert's reward as:
\begin{align}\label{eq:plan}
r^{expert}_t = \frac{1}{|T_{REC}|} \cdot \sum_{t \in T_{REC}} \delta^{t-1} \cdot b_{t}, 
\end{align}
where $\delta$ is a discount factor\footnote{we use $\delta=0.5$.} to
encourage earlier recommendations, 
$b_t$ is the reward obtained at each recommendation made, and
$|T_{REC}|$ is the number of recommendations made. 
$b_t$ is 0 unless the correct movie was recommended.

We define the reward function $\mathcal{R}$ as follows:
\begin{align}
\mathcal{R}\,(x_t) =  \sum_{x_t \in X^{expert} } \gamma^{T-t} (r^{expert}_t - \mu ) 
\end{align}
where $\mu=\frac{1}{t}\sum_{1..t} r^{expert}_t$ is the average of the rewards received by the expert until time $t$ and $\gamma$ is a discount factor to diminish the reward of earlier actions.
We optimize the expected reward for each turn of dialogue $x_t$ and calculate its gradient using REINFORCE \cite{williams1992simple}.
The final role-playing objective $\mathcal{L}_{RP}$ is:
\begin{align}
\nabla\mathcal{L}_{RP}(\theta;\textsc{z}) = \sum_{x_t \in X^{expert} } \mathbb{E}_{x_t} [ \nabla \log p(x_t | x_{<t})  \mathcal{R}\,(x_t) ] 
\end{align}
We optimize the role-playing objective with the pre-trained expert model's decision ($\mathcal{L}_{decide} $) and generation ($\mathcal{L}_{gen}$) objectives at the same time.
To control the variance of the RL loss, we alternate optimizing the RL loss and other two supervised losses for each step.
We do not fine-tune the prediction loss, in order not to degrade the prediction performance during bot-play.

%% file: 04experiment.tex
\section{Experiments}\label{sec:exp_dialrec}

\begin{table*}[h]
\centering
\begin{tabular}{c|r  cc  cccc  c}
\toprule
&&\multicolumn{2}{c}{\textbf{Generation}} & \multicolumn{4}{c}{\textbf{Recommendation}} & \multicolumn{1}{c}{\textbf{Decision}}  \\
\cmidrule(lr){3-4} \cmidrule(lr){5-8} \cmidrule(lr){9-9}
& & \texttt{F1} & \texttt{BLEU}  &\texttt{Turn@1} & \texttt{Turn@3} & \texttt{Chat@1} & \texttt{Chat@3} & \texttt{Acc} \\
\midrule
\parbox[t]{2mm}{\multirow{4}{*}{\rotatebox[origin=c]{90}{{{Baseline}}}}} &
\textsc{TFIDF-Ranker} & 	32.5 &	\textbf{27.8}   & - & - & - & - & - \\
&\textsc{BERT-Ranker} &  38.3 & 23.9 & - & - & - & - &\\
&\textsc{Random Recc.} &	3.6 & 	0.1 & 21.3 &	59.2 &	23.1 &	62.2  & - \\
&\textsc{BERT Recc.} & 	16.5 &	0.2 & 25.5 &	66.3 &	26.4 &	68.3 & - \\
\midrule
\parbox[t]{2mm}{\multirow{4}{*}{\rotatebox[origin=c]{90}{{{Ours}}}}} 
&\textsc{Generate} 	&39.5&	26.0 &  - & - & - & - & - \\
&\textsc{+predict} 	&40.2&	26.4  & 76.4 &	96.9 &	75.7 &	97.0& - \\
&\textsc{+Decide} &	\textbf{41.0} &	27.4  & \textbf{77.8} &	\textbf{97.1} &	\textbf{78.2} &	\textbf{97.7} & \textbf{67.6} \\
&\textsc{+Plan} & 	40.9 &	26.8 &  76.3 &	95.7 &	77.5 &	97.6  & 53.6 \\
\bottomrule
\end{tabular}
\caption{\label{tab:performance} Evaluation on supervised models. We incrementally add different aspects of modules: \textsc{Generate}, \textsc{predict}, and \textsc{Decide} for supervised multi-aspect learning and \textsc{Plan} for bot-play fine-tuning.
}
\end{table*}

We describe our experimental setup in \S\ref{sec:setup}.
We then evaluate our supervised and unsupervised models in \S\ref{sec:eval_supervised} and \S\ref{sec:eval_game}, respectively.

\subsection{Setup}\label{sec:setup}

We select 5\% of the training corpus as validation set in our training.


All hyper-parameters are chosen by sweeping different combinations and choosing the ones that perform best on the validation set. In the following, the values used for the sweep are given in brackets.
Tokens of textual inputs are lower-cased and tokenized using byte-pair-encoding (BPE) \cite{sennrich2015neural} or the  Spacy\footnote{\url{https://spacy.io/}} tokenizer. 
The seq-to-seq model uses 300-dimensional word embeddings initialized with GloVe \cite{pennington2014glove} or Fasttext \cite{joulin2016bag} embeddings, $[1,2]$ layers of $[256,512]$-dimensional Uni/Bi-directional LSTMs \cite{hochreiter1997long} with 0.1 dropout ratio, and soft attention \cite{Bahdanau2015NeuralMT}.
At decoding, we use beam search with a beam of size 3, and choose the maximum likelihood output.
For each turn, the initial movie text and all previous dialogue turns including seeker's and expert's replies are concatenated as input to the models.

Both supervised and bot-play learning use Adam \cite{Kingma2015AdamAM} optimizer with batch size 32 and learning rates of $[0.1,0.01,0.001]$ with 0.1 gradient clipping. 
The number of softmax layers \cite{yang2017breaking} is $[1, 2]$.
For each turn, the initial movie description and all previous dialogue utterances from the seeker and the expert are concatenated as input text to the other modules. 
Each movie textual description is truncated at 50 words for efficient memory computation.

We use annealing to balance the different supervised objectives: we only optimize the \textsc{generate} loss for the first 5 epochs, and then gradually increase weights for the \textsc{predict} and \textsc{decide} losses. 
We use the same movie-sets as in the supervised phase to fine-tune the expert model.
Our models are implemented using PyTorch and ParlAI \cite{miller2017parlai}.
Code and dataset will be made publicly available through ParlAI\footnote{\url{https://github.com/facebookresearch/ParlAI}}.

\subsection{Evaluation of Supervised Models}\label{sec:eval_supervised}
\paragraph{Metrics.}
We first evaluate our supervised models on the three supervised tasks: dialogue generation,  movie recommendation, and per-turn decision to speak or recommend. 
The dialogue generation is evaluated 
using the F1 score and BLEU \cite{papineni2002bleu} comparing the predicted and ground-truth utterances.
The F1 score is computed at token-level.
The recommendation model is evaluated by calculating the percentage of times the correct movie is among the top k recommendations (hit@k).
In order to see the usefulness of dialogue for recommendation, precision is measured per each expert turn of the dialogue (\texttt{Turn@k}) regardless of the decision to speak or recommend, and at the end of the dialogue (\texttt{Chat@k}).  

\paragraph{Models.}
We compare our models with Information Retrieval (IR) based models and recommendation-only models. 
The IR models retrieve the most relevant utterances from the set of candidate responses of the training data and rank them by comparing cosine similarities using TFIDF features or BERT \cite{devlin2018bert} encoder features. 
Note that IR models make no recommendation.
The recommendation-only models always produce recommendation utterances following the template (e.g., ``how about this movie, \texttt{[MOVIE]}?'') where the \texttt{[MOVIE]} is chosen randomly or based on cosine similarities between dialogue contexts and the text descriptions of candidate  movies. 
We use the pre-trained BERT encoder \cite{devlin2018bert} to encode dialogue contexts and movie text descriptions.

We incrementally add each module to our base \textsc{Generate} model: \textsc{Predict} and \textsc{Decide} for supervised learning and \textsc{Plan} for bot-play fine-tuning.
Each model is chosen from the best model in our hyper-parameter sweeping.

\paragraph{Results.}
Table \ref{tab:performance} shows performance comparison on the test set.
Note that only the full supervised model (+\textsc{Decide}) and the fine-tuned model (+\textsc{Plan}) can appropriately operate every function required of an expert agent such as producing utterances, recommending items, and deciding to speak or recommend.

Compared to recommendation-only models, our prediction \textsc{Predict} modules show significant improvements over the recommendation baselines on both per-turn and per-chat recommendations: 52\% on \texttt{Turn@1} and 34\% on \texttt{Turn@3}.
\texttt{Chat} scores are always higher than \texttt{Turn}, indicating that recommendations get better as more dialogue context is provided. 
The \textsc{Decide} module yields additional improvements over the \textsc{Predict} model in both generation and recommendation, with 67.6\% decision accuracy, suggesting that the supervised signal of decisions to speak or recommend can contribute to better overall representations. 

In generation, our proposed models show comparable performance as the IR baseline models (e.g., BERT\textsc{Ranker}).
The +\textsc{Decide} model improves on the F1 generation score because it learns when to predict the templated recommendation utterance.

As expected, +\textsc{Plan} slightly hurts most metrics of supervised evaluation, because it optimizes a different objective (the game objective), which might not systematically align with the supervised metrics. For example, a system optimized to maximize game objective should try to avoid incorrect recommendations even if humans made them.
Game-related evaluations are shown in \S\ref{sec:eval_game}.

\begin{figure}[h!]
\vspace{0mm}
\centering
{
\subfloat[{Rank of recommendation}]{
\includegraphics[trim=8.4cm 0.2cm 0.2cm 0.0cm,clip,valign=b,width=.5\linewidth]{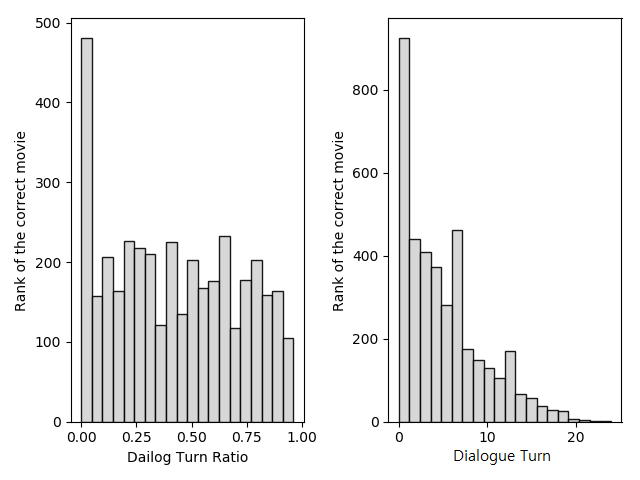}
}
\\
\subfloat[{{\texttt{F1}/\texttt{BLEU} over dialogue turn ratio }}]{
\begin{minipage}[b]{0.99\linewidth}
\hfill
\,\,
\includegraphics[trim=0.2cm 3.2cm 5.74cm 0.2cm,clip,width=.455\linewidth]{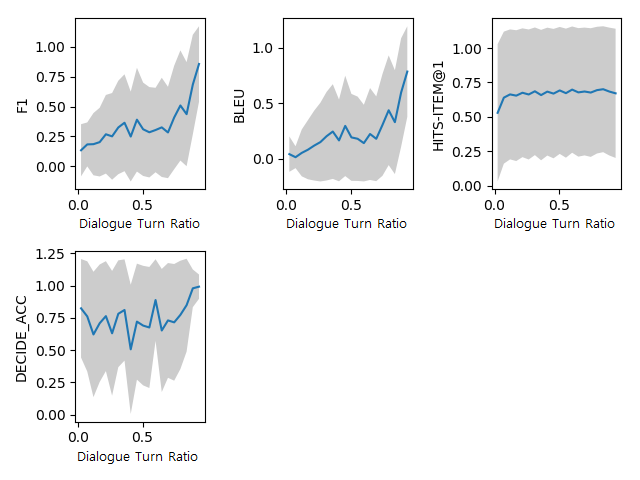}
\,\,
\includegraphics[trim=2.97cm 3.2cm 2.97cm 0.2cm,clip,width=.455\linewidth]{figs/eval_game/{edited_plot_metric_histo_ratio_models_Supervised_hs=512_nl=1_emb=glove_dicttok=bpe_lr=0.001_read=xent_decide=mlp_role=expert.png_binned_ratio}.png}
\,
\end{minipage}
}
\\
\subfloat[{{\texttt{Turn@1} /Decision \texttt{Acc} over dialogue turns}}]{
\begin{minipage}[b]{0.99\linewidth}
\hfill
\includegraphics[trim=5.8cm 0.1cm 5.6cm 6.2cm,clip,width=.465\linewidth]{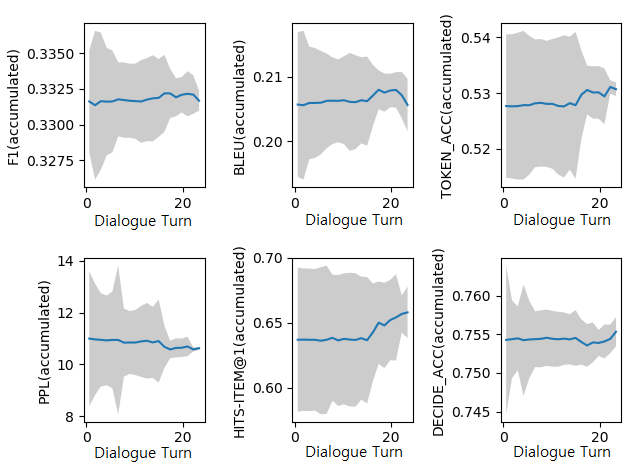}
\@
\includegraphics[trim=11.3cm 0.1cm 0.1cm 6.2cm,clip,width=.465\linewidth]{figs/eval_game/{edited_plot_report_histo_ratio_models_Supervised_hs=512_nl=1_emb=glove_dicttok=bpe_lr=0.001_read=xent_decide=mlp_role=expert.png_binned_ratio.png_binned}.png}
\,
\end{minipage}
}
}
\caption{\label{fig:analysis_eval} Analysis of the expert's model: as the dialogue continues (x-axis is either fraction of the full dialogue, or index of dialogue turn), y-axis is (a) rank of the correct recommendation (the lower rank, the better) and (b,c) \texttt{F1}/\texttt{BLEU}/\texttt{Turn@1}/Decision \texttt{Accuracy} (the higher the better) with the variance shown in grey. 
}
\end{figure}

\paragraph{Analysis}
We analyze how each of the supervised modules acts over the dialogue turns on the test set.
Figure \ref{fig:analysis_eval}(a) shows a histogram of the rank of the ground-truth movie over turns.
The rank of the model's prediction is very high for the first few turns, then steadily decreases as more utterances are exchanged with the seeker.
This indicates that the dialogue context is crucial for finding good recommendations.

The evolution of generation metrics (\texttt{F1}, \texttt{BLEU}) for each turn is shown in Fig.~\ref{fig:analysis_eval}(b), and the (accumulated) recommendation and decision metrics (\texttt{Turn@1}/\texttt{Accuracy}) in Fig.~\ref{fig:analysis_eval}(c)\footnote{For better understanding of the effect of recommendation and decision, we show accumulated values, and per-turn values for generation.}. 
The accumulated recommendation and decision performance sharply rises at the end of the dialogue and variance decreases. 
The generation performance increases, because longer dialogue contexts helps predict the correct utterances.

\subsection{Evaluation on Dialogue Games}\label{sec:eval_game}

\paragraph{Metrics.}
In the bot-play setting, we provide game-specific measures as well as human evaluations.
We use three automatic game measures:
\texttt{Goal} to measure the ratio of dialogue games where the goal is achieved (i.e., recommending the correct movie or not), \texttt{Score} to measure the total game score, and \texttt{Turn2G} to count the number of dialogue turns taken until the goal is achieved.

We conduct human evaluation by making the expert model play with human seekers. We measure automatic metrics as well as dialogue quality scores provided by the player: fluency, consistency, and engagingness (scored between 1 and 5) \cite{zhang2018personalizing}.
We use the full test set (i.e., 911 movie sets) for bot-bot games and use 20 random samples from the test set for \{bot,human\}-human games.

\paragraph{Models.}
We compare our best supervised model with several variants of our fine-tuned bot-play models. We consider bot-play of an expert model with different seeker models such as BERT-Ranker based seeker and Seq-to-Seq based seeker.
Each bot-play model is trained on the same train set that is used for training the original supervised model. The seeker model uses retrieval based on BERT pretrained representations of dialogue context (\texttt{BERT}-\textsc{R}) \footnote{A potential direction for future work may have more solid seeker models and explore which aspect of the model makes the dialogue with the expert model more goal-oriented or human-like.}.

\begin{table}[h]
\small
\centering
\begin{tabularx}{\columnwidth}{@{}
l @{\hskip 1mm} l @{\hskip 2mm} 
@{\hskip 1mm} c @{\hskip 1mm}c @{\hskip 1mm} c @{\hskip 2mm}
@{\hskip 1mm} c @{\hskip 1mm} c @{\hskip 1mm} c@{}}
\toprule
\multicolumn{2}{@{}c@{\hskip 3mm}}{Players} & \multicolumn{3}{@{}c@{\hskip 3mm}}{Automatic} & \multicolumn{3}{@{}c@{}}{Human}\\
\cmidrule(rr){1-2} \cmidrule(rr){3-5} \cmidrule(lr){6-8}
\textbf{Expert} & \textbf{Seeker}& \textbf{\texttt{Goal}} & \textbf{\texttt{Sco}} & \textbf{\texttt{T2G}}& \textbf{\texttt{F}}  &  \textbf{\texttt{C}} &  \textbf{\texttt{E}}\\
\midrule
Supervised$^*$ & \texttt{BERT}-\textsc{R} &30.9	 & 38.3	&  1.4& -& -& -\\
Bot-play \textbackslash w S2S & \texttt{BERT}-\textsc{R} & 42.1& 49.6& 2.8 & -& - & -\\
Bot-play \textbackslash w \texttt{BERT}-\textsc{R} & \texttt{BERT}-\textsc{R} & \textbf{48.6}& \textbf{52.4}& \textbf{3.2}& -& -& - \\
\midrule
Supervised$^*$ &Human & 55.0 & 51.2 & 2.1 &  3.1 & 2.2 & \textbf{2.0} \\
Bot-play$^*$ &Human & \textbf{68.5} & \textbf{54.7} & \textbf{3.1} & \textbf{3.2} & \textbf{2.6} & \textbf{2.0} \\
\midrule
Human &Human & 95.0 & 64.3	&8.5 & 4.8&	4.7&	4.2\\
\bottomrule
\end{tabularx}
\caption{\label{tab:game_eval} Evaluation on dialogue recommendation games: bot-bot (top three rows) and \{bot,human\}-human (bottom three rows). 
We use automatic game measures (\textbf{\texttt{Goal}}, \texttt{\textbf{Sco}re}, \texttt{\textbf{T}urn\textbf{2G}oal}) and human quality ratings (\texttt{\textbf{F}luency}, \texttt{\textbf{C}onsistency}, \texttt{\textbf{E}ngagingness}).
}
\end{table}

\paragraph{Results.}
Compared to the supervised model, the self-supervised model fine-tuned by seeker models shows significant improvements in the game-related measures. 
In particular, the \texttt{BERT}-\textsc{R} model shows a +27.7\% improvement in goal success ratio.
Interestingly, the number of turns to reach the goal increases from 1.4 to 3.2, indicating that conducting longer dialogues seems to be a better strategy to achieve the game goal throughout our role-playing game.

In dialogue games with human seeker players, the bot-play model also outperforms the supervised one, even though it is still far behind human performance. 
When the expert bot plays with the human seeker, performance increases compared to playing with the bot seeker, because the human seeker produces utterances more relevant to their movie preferences, increasing overall game success.

%% file: 06related.tex
\section{Related Work}\label{sec:related_dialrec}

Recommendation systems often rely on matrix factorization \cite{Koren2009MatrixFT,He2017NeuralCF}.
Content \cite{mooney2000content} and social relationship features \cite{ma2011recommender} have also been used to help with the cold-starting problem of new users. The idea of eliciting users' preference for certain content features through dialogue has led to several works.
\citet{Wrnestl2004ModelingAD} studies requirements for developing a conversational recommender system, e.g., accumulation of knowledge about user preferences and database content.
\citet{Reschke2013GeneratingRD} automatically produces template-based questions from user reviews.
However, no conversational recommender systems have been built based on these works due to the lack of a large publicly available corpus of human recommendation behaviors.

Very recently, \citet{li2018towards} collected the \textsc{ReDial} dataset, comprising 10K conversations of movie recommendations, and used it to train a generative encoder-decoder dialogue system.
In this work, crowdsource workers freely talk about movies and are instructed to make a few movie recommendations before accepting one.
Compared to \textsc{ReDial}, our dataset is grounded in real movie preferences (movie ratings from MovieLens), instead of relying on workers' hidden movie tastes. This allows us to make our task goal-directed rather than chit-chat; we can optimize prediction and recommendation strategy based on a known ground truth, and train the \textsc{predict} and \textsc{plan} modules of our system. That in turn allows for novel setups such as bot-play.

To the best of our knowledge, \citet{bordes2016learning} is the only other goal-oriented dialogue benchmark grounded in a database that has been released with a large-scale publicly available dataset. Compared to that work, our database is made of real (not made-up) movies, and the choice of target movies is based on empirical distances between movies and movie features instead of being arbitrary. This, combined with the collaborative set-up, makes it possible to train a model for the seeker in the bot-play setting.

Our recommendation dialogue game is collaborative. Other dialogue settings with shared objectives have been explored, for example a collaborative graph prediction task
\cite{He2017LearningSC}, and semi-cooperative negotiation tasks \cite{Lewis2017DealON,Yarats2018HierarchicalTG,he-etal-2018-decoupling}.

%% file: 05discussion.tex
\section{Conclusion and Future Directions}

In conclusion, we have posed recommendation as a goal-oriented game between an expert and a seeker, and provided a framework for both training agents in a supervised way
by learning to mimic a large set of collected human-human dialogues, as well as by bot-play between trained agents.
We have shown that a combination of the two stages leads to learning better expert recommenders.

Our results suggest several promising directions.
First, we noted that the recommendation performance linearly increases as more dialogue context is provided. An interesting question is how to learn
to produce the best questions that will result in
the most informative dialogue context.
    
Second, as the model becomes better at the game, we observe an increase in the length of dialogue.
However, it remains shorter than the average length of human dialogues, possibly because our reward function is designed to minimize it, which worked better in experiments.
A potential direction for future work is to study how different game objectives interact with each other.

Finally, our evaluation on movie recommendation is made only within the candidate set of movies given to expert.
Future work should evaluate if our training scheme generalizes to a fully open-ended recommendation system, thus making our task not only useful for research and model development, but a useful end-product in itself.




%% file: appendix.tex
\renewcommand*\appendixpagename{\Large Appendices}
\clearpage
\begin{appendix}\label{sec:appendix_dialrec}

\section{Additional Notes on Data Preparation}
we obtain a rating matrix of 265,905 users and 11,382 movies. 
We filter the data according to a few criteria:
\begin{itemize}[noitemsep,topsep=0pt,leftmargin=*]
\item users who watched less than 50 movies are filtered out.
\item moves which are watched less than 50 users are filtered out.
\item movies which are filmed before 1950 are filtered out.
\item movies whose average rates are less than 2 and users who average rates are less than 2 are filtered out.
\end{itemize}

We also remove some movie sets which are too difficult or too easy to predict based on their distance scores.
For example, we filter out movie sets where the cosine similarity of the correct movie and the averaged incorrect movies is less than 0.75. 
After filtering, the remaining data comprises 5,330 movies, rated by 65,181 users.

We tested different types of embedding features such as movie IDs (i.e., MovieLens's ratings), movie text (i.e., Wiki-text), and knowledge base features (e.g., director's name). The movie ID features turn out to be the best performing for recommendation performance. 
After training, the model finds reasonable close neighbors; for example, for ``Ice Age'', the model identifies ``Shrek 2'', ``Shrek'', ``Monsters Inc.'', and ``Finding Nemo'' as close. 

\section{Data Collection: Full Description}
In our annotation interface, we provide action buttons for workers to click on in order to interact with the system. When a button is clicked, the corresponding system message is shown. For example, if an expert clicks on a movie button to recommend that movie, the system displays a recommendation message to the seeker, using a simple template.
Similarly, if a seeker clicks to accept or reject the recommendation, a templated message with the decision is automatically delivered to the expert.

If an expert recommends the correct movie, a seeker accepts the correctly recommended movie, or a seeker rejects an incorrectly recommended movie,  they receive a  reward (points, which can translate into bonus money if enough points are earned); 
otherwise, the system encourages them to focus more on the task and get more points. 
The amount of reward points awarded is calculated based on the similarities between the average of the seeker's movie set and each candidate movie in the expert's set, using a softmax. 
The similarity scores are calculated using the euclidean distance between movie embedding vectors (see Section~\ref{sec:sup_app}).

Overall, a total of 1,034 unique workers created 
9,125 dialogues, over a duration of 2.5 weeks.

\begin{figure*}[h]
\vspace{0mm}
\centering
{
\includegraphics[trim=0cm 4.5cm 8.3cm 0cm,clip,width=.99\linewidth]{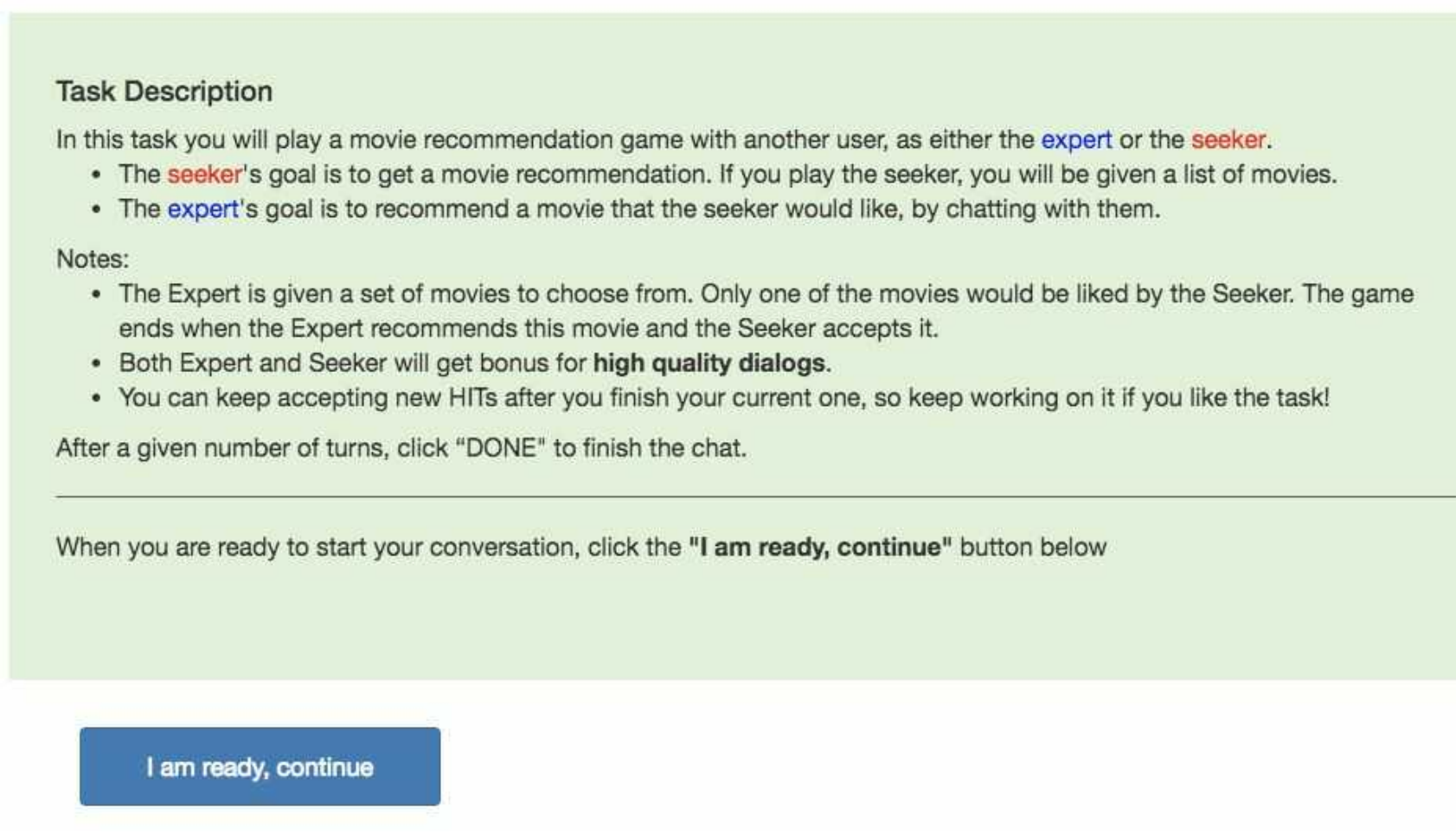}
}
\caption{\label{fig:amt_snapshot_task} Interface of our data collection (1): task description page.
}
\vspace{0mm}
\end{figure*}
\begin{figure*}[h]
\vspace{0mm}
\centering
{
\subfloat[Seeker's left panel]
{
\includegraphics[trim=0cm 0cm 13cm 0cm,clip,width=.5\linewidth]{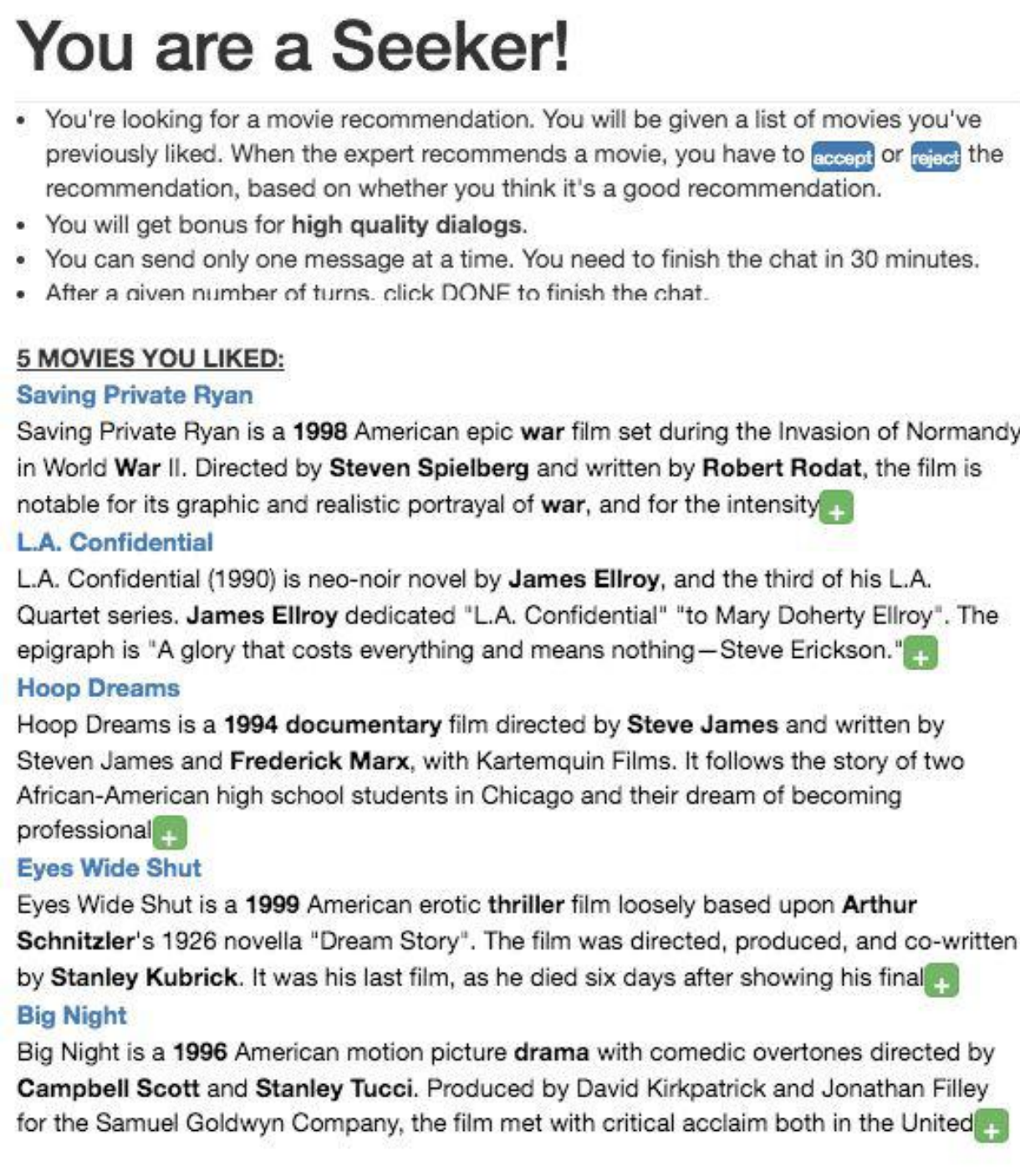}
}
\subfloat[Seeker's right panel]
{
\includegraphics[trim=0cm 0cm 9cm
0cm,clip,width=.5\linewidth]{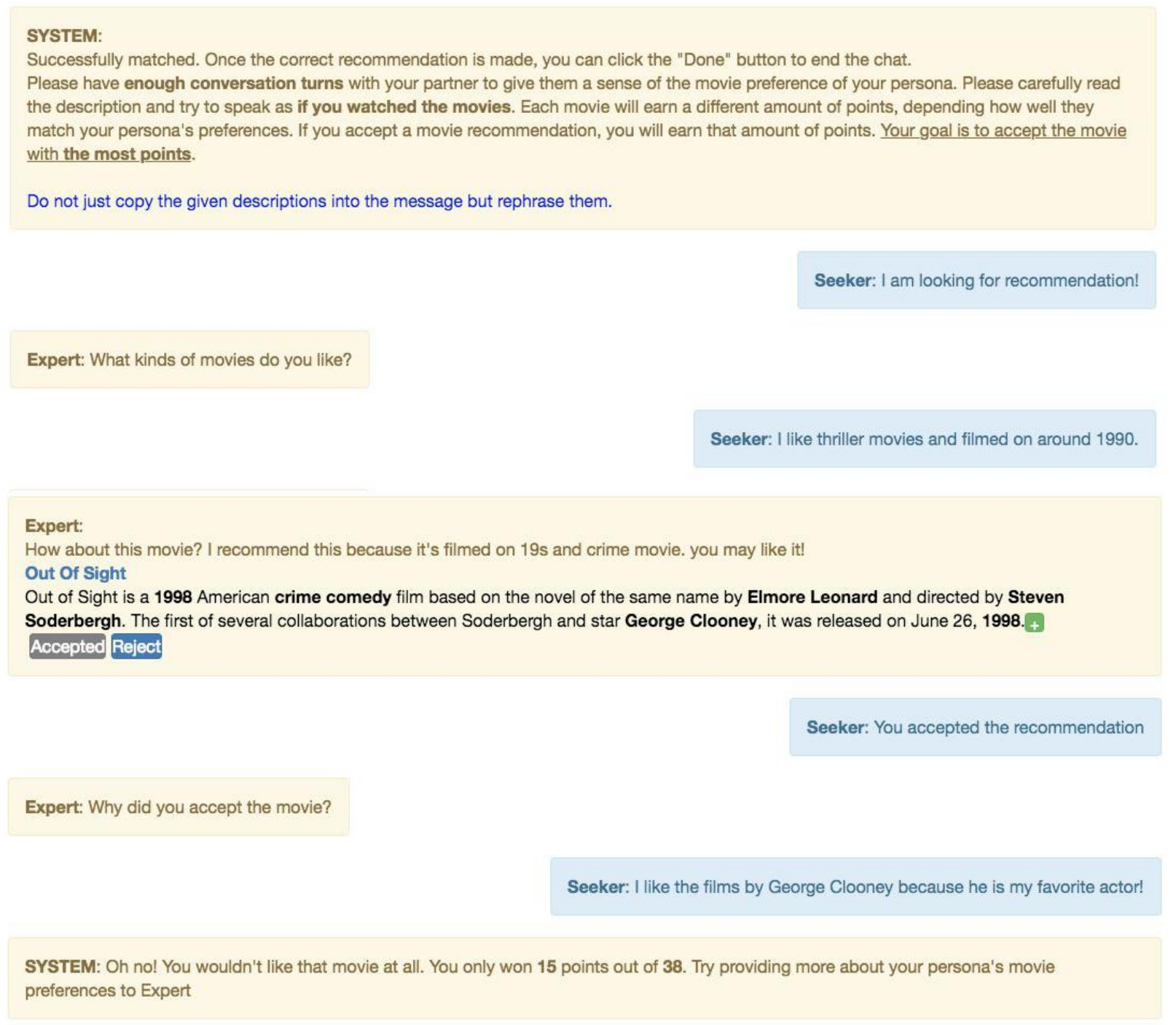}
}
\\
\subfloat[Expert's left panel]
{
\includegraphics[trim=0cm 0cm 13cm 0cm,clip,width=.5\linewidth]{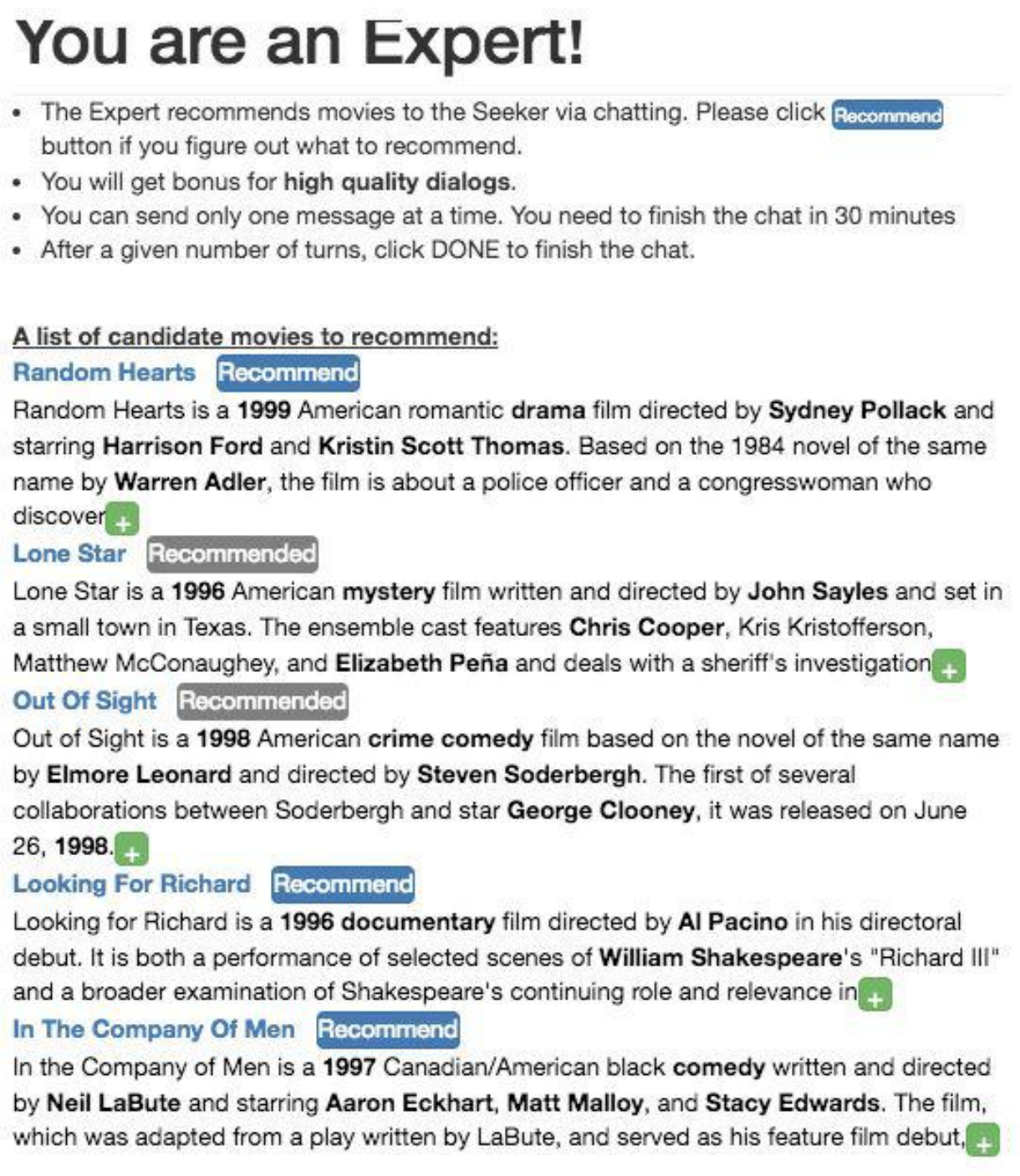}
}
\subfloat[Expert's right panel]
{
\includegraphics[trim=0cm 0cm 12cm
0cm,clip,width=.5\linewidth]{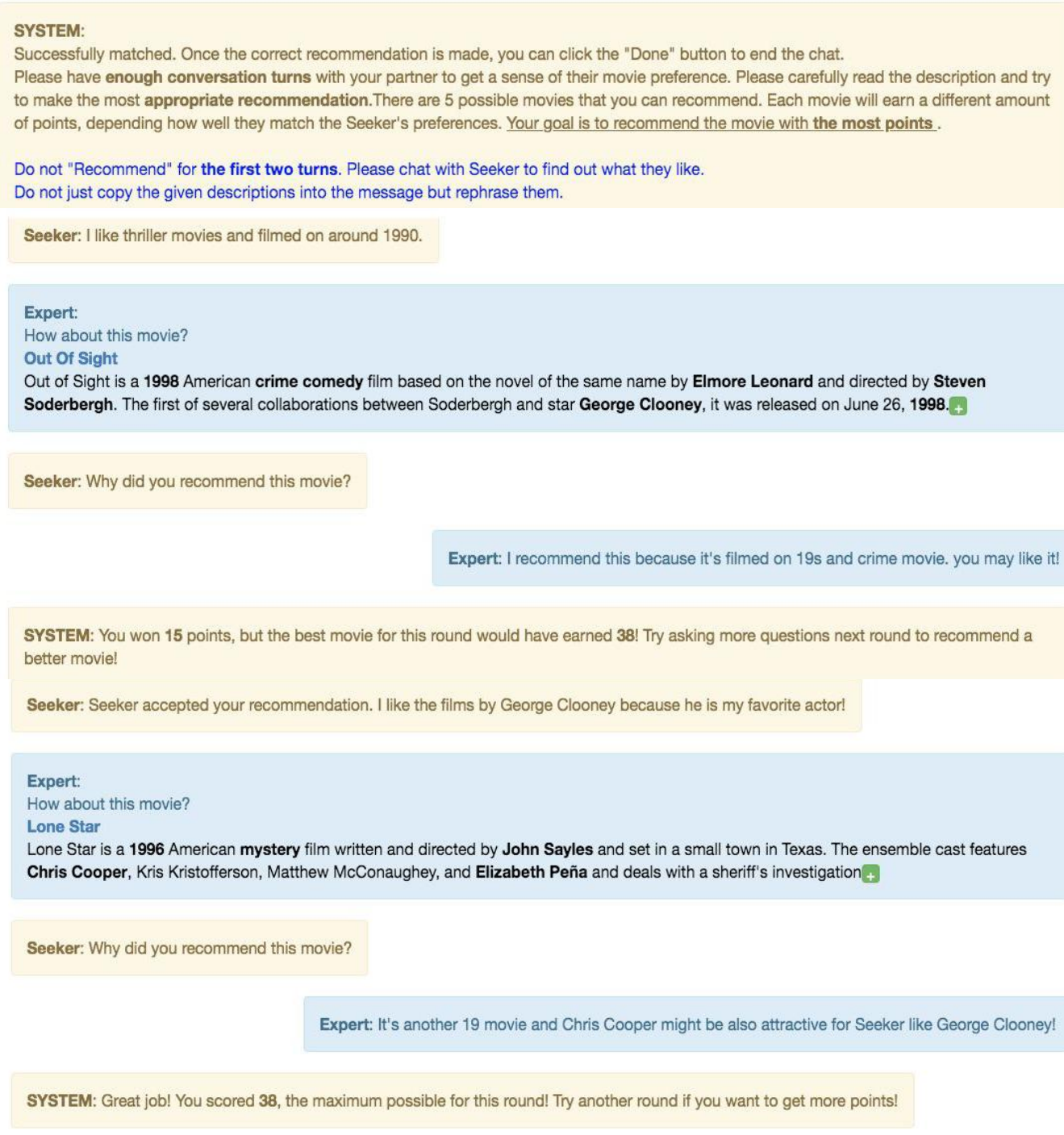}
}
}
\caption{\label{fig:amt_snapshot} Interface of our data collection (2): seeker's and expert's pages.
}
\vspace{0mm}
\end{figure*}


\section{Supervised training: Details}
\label{sec:sup_app}
This section gives more details about the supervised training phase.

\paragraph{Encoding textual inputs:} Textual inputs are encoded differently for the dialogue context and for the movie descriptions. The dialogue history context $h_t$ for predicting utterance $x_{t+1}$ comprises the history of all previous utterances $x_1,\cdots,x_t$. Each utterance is encoded with an LSTM \cite{hochreiter1997long}. The dialogue context is then obtained by averaging over all utterances:
\begin{align}\label{eq:ht_app}
h_t &= \texttt{AVG}\,(\texttt{LSTM}(\texttt{x}_1),\ldots,\texttt{LSTM}(\texttt{x}_t))
\end{align}

For the movies, we found that using bags of words instead worked better. We encode each sentence of a movie description as a bag of words, and then average all the resulting representations to obtain $m_j$, the representation of the $j$-th movie:
\begin{align}\label{eq:movies_app}
\textsce{E}(\texttt{m}_j) &= \texttt{AVG}\,(\texttt{BOW}(\texttt{m}_j)) \quad\quad \texttt{for} j\in1..K
\end{align}

\paragraph{Aligning dialogue context and movie descriptions:}
we use dot-product attention\cite{chen2017reading} between the dialogue context and each of the movie descriptions:
\begin{align}\label{eq:attention_app}
\texttt{c}_j  &=  h_t \boldsymbol{\cdot} m_j \quad\quad \texttt{for} j\in1..K
\end{align}

\paragraph{Generating utterances: \textsc{Generate}}
The expert can produce two types of utterances, according to whether it is recommending a movie or asking for more input from the seeker.
For \texttt{Recommend}, the response is produced by a template: ``How about this movie, \texttt{[MOVIE]}?'' where \texttt{[MOVIE]} is the movie that the expert is recommending. 
For \texttt{Speak}, the next utterance is generated by taking the dialogue context history $h_t$ and the average of all movie representations $\texttt{M} = \texttt{AVG}(m_1,..,m_K)$, and inputting them into a seq2seq generative model with attention \cite{Bahdanau2015NeuralMT}.
The model is then trained to minimize the negative log likelihood of the true next utterance $x_{t+1}$ according to the model distribution $p_{gen}$:
\begin{align}\label{eq:generatesuppl}
\mathcal{L}_{gen} &= -\log p_{gen}(x_{t+1} | h_t, \texttt{M}), where\\
\texttt{M} &= \texttt{AVG}(m_1,..,m_K)
\end{align}
We \textit{include} \texttt{Recommend} utterances in the $\mathcal{L}_{gen}$ calculation; as a result, the generation loss is also a partial indicator of other aspects such as \textsc{Decide} and \textsc{Predict}, in addition to the corresponding specific losses (see below).

\paragraph{Predicting the correct movie to recommend: \textsc{Predict}} Let $y$ denote the correct movie. The prediction module is trained by minimizing the negative log likelihood of $y$ according to the distribution of a softmax predictor over the $\texttt{c}_j$ inputs described above:

\begin{align}\label{eq:predict_app}
\mathcal{L}_{predict} &= -\log p(y | c_1, \cdots , c_K),\quad\text{where}\\
\texttt{c}_j  &=  h_t \boldsymbol{\cdot} m_j \quad\quad \text{for}~ j\in1..K
\end{align}

When making a recommendation, the expert recommends the top candidate: $\arg\max_c \{\texttt{r}_1..\texttt{r}_K\}$.
We also experimented with using a soft representation for the target movie distribution, for example through a softmax over similarities.
For instance, in Figure \ref{fig:example_dialog_long}, the hard ground-truth movie distribution is $\{1,0,0,0,0\}$, and the soft version is $\{0.37,0.15,0.16,0.16,0.15\}$. But  the hard version always outperformed the soft version in our experiments.

\paragraph{Deciding when to recommend: \textsc{Decide}} The expert needs to decide whether to to recommend a movie or speak to elicit more information. We model this using a two-layer perceptron that takes the movie prediction distribution scores and the dialogue context as input, and predicts whether to make a recommendation or not. Training is conducted by minimizing the negative log likelihood of the ground truth decision:
\begin{align}\label{eq:decidesuppl}
\mathcal{L}_{decide} &= p_{MLP}(d_{t+1}|h_t,c_1, \cdots, c_K )
\end{align}

We also experimented with other functions of the movie prediction distribution (e.g., skewness and kurtosis \cite{mardia1970measures}), but the multi-layer perceptron (\texttt{MLP}) always performed better.

\paragraph{Supervised loss of the overall system:} The overall objective function of the full supervised system is as follows:
\begin{align}\label{eq:supervisedsuppl}
\mathcal{L}_{sup} = 
\alpha\mathcal{L}_{gen} + \beta\mathcal{L}_{predict} + (1{-}\alpha{-}\beta)\mathcal{L}_{decide} 
\end{align}
where $\alpha$ and $\beta$ are weight terms that control the balance between the different objectives and are empirically optimized on the validation set.
For the \textsc{predict} and \textsc{decide} losses, we use annealing at the beginning of training, with all the weight being given to the \textsc{generate} loss, and the weights of the other two being gradually increased.

\end{appendix}